# Elagage d'un perceptron multicouches : utilisation de l'analyse de la variance de la sensibilité des paramètres.

Philippe Thomas, Andre Thomas

Centre de Recherche en Automatique de Nancy CRAN-UMR 7039), Nancy-Université, CNRS

ENSTIB 27 rue du merle blanc, B.P. 1041

88051 Epinal cedex 9 France

philippe.thomas@cran.uhp-nancy.fr

*Résumé*— Une phase primordiale lors de la modélisation d'un système à l'aide d'un perceptron multicouches reste la détermination de la structure du réseau. Dans ce cadre, nous proposons un algorithme d'élagage du réseau basé sur l'utilisation de l'analyse de la variance de la sensibilité de tous les paramètres du réseau. Cet algorithme sera testé sur deux exemples de simulation et ses performances seront comparées à trois autres algorithmes d'élagage de la littérature.

*Mots clés*— perceptron multicouches, réseau de neurones, élagage, sélection de structure.

## I. Introduction

Depuis les premiers travaux de Rumelhart et McCleland [26] qui sont à l'origine des réseaux de neurones et, plus particulièrement des perceptrons multicouches, nombre d'applications les ont utilisés dans des domaines divers. Cependant, quelque soit le domaine d'application, la détermination de la structure reste un problème crucial lors de la phase de modélisation. En effet, même si les travaux de Cybenko [7] et Funahashi [13] ont montré qu'une seule couche cachée utilisant des fonctions d'activation du type sigmoïdal était suffisant pour pouvoir approximer toute fonction non linéaire avec la précision voulue, rien n'est dit a priori sur le nombre de neurones cachés à utiliser. L'idée la plus naturelle consisterait donc à choisir le plus grand nombre de neurones cachés possible ce qui permettrait d'obtenir la plus grande précision. Malheureusement, nous ne nous posons pas un problème d'approximation uniforme mais bien un problème d'ajustement d'une fonction à un nombre fini de point [9]. Dans ce cadre, le risque est que le réseau apprenne le bruit sur la fonction et non plus la fonction et nous nous trouvons face à un problème de surapprentissage. Pour répondre à ce problème, diverses techniques ont été proposées comme les méthodes de régularisation telle que l'arrêt prématuré (early stopping) [10] ou les méthodes de pénalisation [1 ; 3]. D'autres auteurs ont proposés de construire de manière itérative la couche cachée [4 ; 12 ;17 ; 22 ; 24 ; 27]. La dernière approche a consisté à partir d'une structure incluant un grand nombre de neurones cachés puis d'éliminer ces derniers en commençant par les moins significatifs [6 ; 15 ; 19 ; 20 ; 23 ; 31 ; 32 ; 33 ; 34]. D'autre part, comme pour tout problème de modélisation, la sélection des variables d'entrée est une tâche primordiale et il est indispensable que l'ensemble des variables d'entrée soit aussi réduit que possible afin de limiter le nombre de paramètres du modèle, toujours dans le but d'éviter les phénomènes de surapprentissage. Plusieurs techniques de sélection de variables basées sur les statistiques notamment ont été proposées. On peut citer, par exemple, l'analyse en composante principale [16], l'analyse en composante curviligne [8] ou encore la méthode du descripteur sonde [9]. D'autres méthodes ont été proposées pour sélectionner les variables dans le cadre strict des réseaux de neurones [2 ; 5 ; 14 ; 18]. Peu de ces méthodes permettent de simultanément sélectionner les variables et d'éliminer les paramètres superflus. Nous pouvons citer entre autres [11 ; 15 ; 28]. L'objectif de ce travail est de proposer une nouvelle méthode permettant simultanément de sélectionner les variables pertinentes tout en supprimant les paramètres superflus du modèle neuronal. Cette méthode est une évolution de l'algorithme proposé par Engelbrecht [11] qui utilise l'analyse de la variance de la sensibilité des neurones cachés et des entrées pour déterminer lesquels doivent être supprimés du modèle. Nous proposons de modifier cet algorithme de manière à pouvoir sélectionner plus finement les paramètres qui doivent être éliminés. Dans un premier temps, nous rappellerons très succinctement la structure et les notations du perceptron multicouches utilisées. Nous présenterons ensuite l'algorithme proposé par Engelbrecht [11] sur lequel repose notre méthode ainsi que l'évolution que nous en proposons. Dans une quatrième partie, nous allons décrire brièvement les algorithmes OBS [15] et N2PFA [28] qui nous serviront de base de comparaison. Enfin, nous finirons en présentant les deux exemples de simulation sur lesquels les quatre algorithmes présentés seront testés et comparés.

## II. Le Perceptron multicouches

Nous allons ici rappeler la structure du perceptron multicouches utilisé. Le réseau présenté par la figure 1 est composé de neurones interconnectés en trois couches successives. La première couche est composée de neurones « transparents » qui n'effectuent aucun calcul mais simplement distribuent leurs entrées à tous les neurones de la couche suivante appelée couche cachée. Les neurones de la couche cachée (figure 1), dont un exemple peut être représenté par la figure 2, reçois les $n_0$ entrées $\{x_1^0, \cdots, x_{n_0}^0\}$ de la couche d'entrée avec les poids associés $\{w_{i1}^0, \cdots, w_{in_0}^0\}$. Ce neurone commence par calculer la somme pondéré de ses $n_0$ entrées :

$$z_i^1 = \sum_{h=1}^{n_0} w_{ih}^1 . x_h^0 + b_i^1 \qquad (1)$$

où $b_i^1$ est un biais (ou seuil).

La sortie du neurone caché est obtenue en transformant la somme (1) par l'intermédiaire de la fonction d'activation g(.) :

$$x_i^1 = g(z_i^1). \tag{2}$$

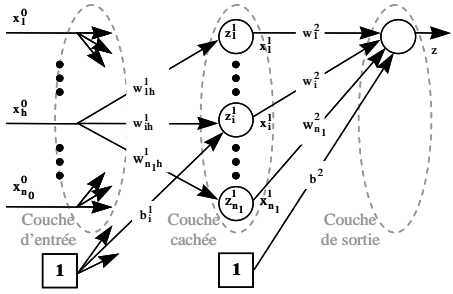

Figure 1. L'architecture du perceptron multicouches

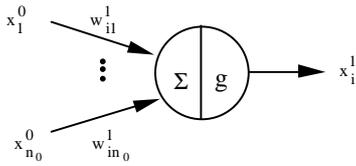

Figure 2. Le neurone i de la couche cachée

Bien que de nombreuses fonctions d'activations ait été proposées, la fonction g(.) est généralement la tangente hyperbolique [32] :

$$g(x) = \frac{2}{1+e^{-2x}} - 1 = \frac{1-e^{-2x}}{1+e^{-2x}}. \tag{3}$$

Le neurone de la dernière couche (ou couche de sortie) utilise une fonction d'activation linéaire et n'effectue donc qu'une simple somme pondérée de ses entrées :

$$z = \sum_{i=1}^{n_1} w_i^2 . x_i^1 + b \tag{4}$$

où $w_i^2$ sont les poids connectant les sorties des neurones cachés au neurone de sortie et b est le biais du neurone de sortie.

## III. L'ALGORITHME

### A. L'algorithme de Engelbrecht [11]

L'algorithme proposé par Engelbrecht repose sur la « variance nullity measure » (VNM) [25 ; 29]. L'idée de base est de tester si la variance de la sensibilité d'une entrée ou de la sortie d'un neurone caché pour différentes données est significativement différente de zéro. Si cette variance n'est pas significativement différente de zéro et si la sensibilité moyenne est petite, alors cela indique que l'entrée ou le neurone caché correspondant a un faible impact ou pas d'impact du tout sur la sortie du réseau considéré. Un test d'hypothèse peut donc utiliser cette VNM pour déterminer statistiquement si un neurone caché ou une entrée doit être éliminé en utilisant une distribution du $\chi^2$. Pour tester si il faut supprimer un neurone caché, nous devons déterminer la VNM du poids $w_i^2$ (i=1…$n_1$) connectant ce neurone caché au neurone de sortie. Pour cela, il est nécessaire de connaître la sensibilité de la sortie z par rapport au paramètre $w_i^2$ et cette sensibilité correspond à la contribution de ce paramètre sur l'erreur totale faite en sortie. Cette contribution est classiquement déterminée par la dérivée partielle de la sortie du réseau z par rapport au paramètre $w_i^2$ considéré :

$$S_{w_i^2}(p) = \frac{\partial z(p)}{\partial w_i^2} = x_i^1(p) \qquad p = 1 \cdots P \tag{5}$$

où p est l'indice de la donnée extraite de la base de données d'apprentissage qui comprend P données.

D'une manière similaire, on peut déterminer la sensibilité de la sortie z par rapport à l'entrée $x_h^0$ (h=1…$n_0$) en effectuant la dérivé partielle de la sortie par rapport à l'entrée $x_h^0$ considérée :

$$\begin{aligned} S_{x_h^0}(p) &= \frac{\partial z(p)}{\partial x_h^0(p)} = \sum_{i=1}^{n_1} \frac{\partial z(p)}{\partial x_i^1(p)} \cdot \frac{\partial x_i^1(p)}{\partial z_i^1(p)} \cdot \frac{\partial z_i^1(p)}{\partial x_h^0(p)} \\ &= \sum_{i=1}^{n_1} w_i^2 \cdot g'(z_i^1(p)) \cdot w_{ih}^1 \\ &= \sum_{i=1}^{n_1} w_i^2 \cdot \left(1 - (x_i^1(p))^2\right) \cdot w_{ih}^1 \qquad p = 1 \cdots P \end{aligned} \tag{6}$$

Nous pouvons noter la sensibilité de la sortie par rapport à un neurone caché ou à une entrée par une notation commune $S_{\theta_k}(p)$ (p=1…P et k=1…K=$n_0+n_1$) avec $\theta_k$ correspondant à $x_h^0$ si on considère l'entrée h et à $w_i^2$ si on considère le neurone caché i et correspondra donc aux équations (6) et (5) respectivement.

La VNM correspond à la variance $\sigma_{\theta_k}^2$ du paramètre $\theta_k$ considéré :

$$\sigma_{\theta_k}^2 = \frac{\sum_{p=1}^{P}\left(S_{\theta_k}(p) - \overline{S_{\theta_k}}\right)^2}{P-1} \qquad p = 1...P \tag{7}$$

où $\overline{S_{\theta_k}}$ est la moyenne de la sensibilité de la sortie à $\theta_k$ :

$$\overline{S_{\theta_k}} = \frac{\sum_{p=1}^{P} S_{\theta_k}(p)}{P} \tag{8}$$

Il reste à tester l'hypothèse que la variance de la sensibilité pour le neurone caché ou pour l'entrée considéré est approximativement nulle. Pour cela, le test d'hypothèse suivant est construit :

$$\mathcal{H}_0: \quad \sigma_{\theta_k}^2 < \sigma_0^2 \tag{9}$$

où $\sigma_0^2$ est un petit réel positif.

En utilisant le fait que la VNM suit une distribution du $\chi^2(\nu)$ à $\nu$=P-1 degrés de liberté dans le cas d'un jeu de données comprenant P données, la réalisation du test (9) s'effectue en comparant la relation :

$$\Gamma_{\theta_k} = \frac{(P-1).\sigma_{\theta_k}^2}{\sigma_0^2} \tag{10}$$

à un seuil critique $\Gamma_c$ obtenu à l'aide des tables du $\chi^2$ :

$$\Gamma_c = \chi^2(\nu, 1-\alpha) \tag{11}$$

où $\nu=P-1$ est le nombre de degrés de liberté et $\alpha$ est le degré de confiance du test.

Si $\Gamma_{\theta_k}$ est supérieur au seuil $\Gamma_c$, le neurone caché ou l'entrée considéré est éliminé. La valeur du paramètre $\sigma_0^2$ est cruciale pour le succès de l'algorithme. Si cette valeur est trop petite, rien ne sera élagué, si elle est trop grande, on risque d'éliminer des neurones cachés ou des entrées significatifs. Aussi, l'algorithme démarre avec une petite valeur de $\sigma_0^2$ (0,001) et multiplie cette valeur par 10 si rien n'est élagué jusqu'à atteindre 0,1 [11].

Dans la suite de ce papier, cet algorithme sera dénommé « Engel ».

### B. La modification proposée

L'algorithme précédemment présenté a pour but d'éliminer les neurones cachés superflus tout en effectuant simultanément la sélection de variables. Cependant, en ce qui concerne l'élimination des variables, cet algorithme est assez monolithique puisque soit la variable d'entrée doit être supprimée du réseau, soit elle doit être conservée auquel cas elle sera distribuée à tous les neurones cachés sans distinction.

Cependant, il nous semble qu'une variable d'entrée peut être utile dans l'évaluation de la sortie d'un neurone caché et être, dans ce même réseau superflue pour un autre neurone caché. Face à une telle situation, l'algorithme Engel risque, soit d'éliminer une variable qui serait partiellement utilisée, soit, de la conserver dans tous les neurones cachés, et donc de conserver des paramètres superflus qui peuvent être cause de perturbations et de surapprentissage.

C'est pourquoi nous proposons de ne plus travailler entrée par entrée mais paramètre par paramètre.

Nous ne nous trouvons donc plus en présence de deux catégories d'éléments à éliminer $\theta_k$ (neurones cachés et entrées) mais de trois :

- poids connectant les entrées aux neurones cachés $w_{ih}^1$,
- biais des neurones cachés $b_i^1$,
- poids connectant les neurones cachés à la sortie $w_i^2$.

Il est à noter que la sensibilité de la sortie z par rapport au biais b du neurone de sortie est constante et vaut 1 et donc, cette approche ne peut pas être utilisée pour l'éliminer si besoin.

Pour chacune de ces catégories de paramètres, il importe donc de donner la sensibilité de la sortie z par rapport aux paramètres.

Pour les poids $w_i^2$ connectant les neurones cachés au neurone de sortie, cette sensibilité est donnée par l'équation (5).

Pour les biais $b_i^1$ des neurones cachés, cette sensibilité correspond à la contribution de ce paramètre sur l'erreur totale faite en sortie. Cette contribution est classiquement déterminée par la dérivée partielle de la sortie du réseau z par rapport au paramètre $b_i^1$ considéré :

$$S_{b_i^1}(p) = \frac{\partial z(p)}{\partial b_i^1} = \frac{\partial z(p)}{\partial x_i^1(p)} \cdot \frac{\partial x_i^1(p)}{\partial z_i^1(p)} \cdot \frac{\partial z_i^1(p)}{\partial b_i^1}$$
$$= w_i^2 \cdot g'\left(z_i^1(p)\right) \cdot 1 \quad (12)$$
$$= w_i^2 \cdot \left(1 - \left(x_i^1(p)\right)^2\right) \quad p = 1 \cdots P$$

D'une manière similaire, on obtient la sensibilité pour les poids $w_{ih}^1$ connectant les entrées aux neurones cachés :

$$S_{w_{ih}^1}(p) = \frac{\partial z(p)}{\partial w_{ih}^1} = \frac{\partial z(p)}{\partial x_i^1(p)} \cdot \frac{\partial x_i^1(p)}{\partial z_i^1(p)} \cdot \frac{\partial z_i^1(p)}{\partial w_{ih}^1}$$
$$= w_i^2 \cdot g'\left(z_i^1(p)\right) x_h^0(p) \quad (13)$$
$$= w_i^2 \cdot \left(1 - \left(x_i^1(p)\right)^2\right) \cdot x_h^0(p) \quad p = 1 \cdots P$$

Tout comme précédemment, nous pouvons noter la sensibilité de la sortie par rapport à un paramètre $S_{\theta_k}(p)$ (p=1…P et k=1…K=$(n_0+2).n_1$) avec $\theta_k$ correspondant à $w_{ih}^1$ si on considère le poids connectant l'entrée h au neurone caché i, à $b_i^1$ si on considère le biais du neurone caché i et à $w_i^2$ si on considère le poids connectant le neurone caché i au neurone de sortie. $\theta_k$ correspondra donc aux équations (13), (12) et (5) respectivement. Une fois ceci posé, le reste de l'algorithme reste identique. Il faut toujours effectué un test d'hypothèse décrit par l'équation (9) qui conduit à la comparaison de la valeur décrite par (10) avec le seuil (11), la détermination de (10) nécessitant toujours le calcul de la VNM par les équations (7) et (8). Le choix des valeurs pour le degré de confiance $\alpha$ du test et du paramètre $\sigma_0^2$ seront les mêmes que pour l'algorithme Engel afin de faciliter la comparaison.

Dans la suite de ce papier, cet algorithme sera dénommé « Engel_mod ».

## IV. LES ALGORITHMES DE COMPARAISON

### A. Optimal Brain Surgeon (OBS) [15]

L'algorithme OBS minimise la sensibilité d'un critère d'erreur sous la contrainte de nullité des paramètres ce qui correspond à la suppression de ce paramètre.

Le critère d'erreur considéré est généralement le critère quadratique :

$$V(\theta) = \frac{1}{P} \sum_{p=1}^{P} (y(p) - z(p,\theta))^2 \quad (14)$$

où $\theta$ est un vecteur regroupant tous les poids et biais du réseau, z est la sortie du réseau et y la sortie désirée.

La sensibilité $\delta V(\theta)$ du critère $V(\theta)$ est approchée à l'aide d'un développement de Taylor autour de $\theta$ à l'ordre 2 :

$$\delta V(\theta) \approx \delta \theta^T V'(\theta) + \frac{1}{2} \delta \theta^T H \delta \theta \quad (15)$$

Considérant que le gradient $V'(\theta)$ s'annule après convergence le premier terme de (15) disparaît ce qui conduit à :

$$\delta V(\theta) = \frac{1}{2} \delta \theta^T H \delta \theta \quad (16)$$

qui implique uniquement le Hessien H. En notant $e_q$ un vecteur canonique sélectionnant le $q^{\text{ème}}$ élément de $\theta$ ($e_q^T = [0\cdots 010\cdots 0]$), la suppression du paramètre $\theta_q$ (i.e. $e_q^T(\delta\theta + \theta) = 0$) doit conduire à un accroissement minimal du critère (14). Aussi, nous pouvons écrire le Lagrangien :

$$\mathcal{L}(\delta\theta) = \frac{1}{2}\delta\theta^T H \delta\theta + \lambda(e_q^T(\delta\theta + \theta)) \quad (17)$$

et la minimisation de ce Lagrangien conduit à la variation du vecteur des paramètres :

$$\delta\theta = -\frac{\theta_q}{H_{qq}^{-1}}H^{-1}e_q \quad (18)$$

où $H_{qq}^{-1}$ est le $q^{\text{ème}}$ terme diagonal de $H^{-1}$. En pratique, le poids à supprimer est celui qui minimise (15). L'équation (18) permet d'effectuer la suppression du paramètre considéré et de distribuer son influence sur les autres paramètres sans avoir à effectuer un nouvel apprentissage. Diverses évolutions de cet algorithme ont été proposées.

La difficulté de cet algorithme consiste à choisir la structure optimale à utiliser puisque aucun critère d'arrêt n'est inclus. Divers critères peuvent être utilisés. Nous en retiendrons deux, la somme quadratique moyenne de l'erreur obtenue sur le jeu de validation (MSSE) et le critère d'erreur de prédiction final (FPE) qui s'évalue sur le jeu d'apprentissage et qui prend en compte le nombre de paramètres [21]. Nous avons utilisé l'algorithme programmé par Norgaard [23] qui permet d'effectuer des réapprentissages entre chaque suppression de poids. Nous présenterons donc les résultats de cet algorithme avec et sans réapprentissage. Nous aurons donc quatre dénominations pour cet algorithme, « OBS_L_FPE » et « OBS_L_MSSE » lorsqu'un réapprentissage sera utilisé avec les critères de sélection FPE et MSSE (respectivement) et « OBS_WL_FPE » et « OBS_WL_MSSE » lorsque les critères de sélection FPE et MSSE (respectivement) seront utilisés sans réapprentissage.

*B. N2PFA [28]*

L'algorithme N2PFA (Neural Network Pruning for Function Approximation) s'appuie sur l'utilisation de l'erreur absolue moyenne (MAD) afin de mesurer la performance du réseau. En particulier, $MAD_T$, calculée sur le jeu d'apprentissage, et $MAD_V$, calculée sur le jeu de validation sont utilisées pour déterminer l'arrêt de la phase d'élagage :

$$MAD_T = M_T = \frac{1}{P_T}\cdot\sum_{p=1}^{P_T}|y_T(p) - z(p)|$$
$$MAD_V = M_V = \frac{1}{P_V}\cdot\sum_{p=1}^{P_V}|y_V(p) - z(p)| \quad (19)$$

où $P_T$ (resp. $P_V$), $y_T$ (resp. $y_V$) sont le nombre de données et les données du jeu d'apprentissage (resp. de validation).

L'algorithme N2PFA, comme les autres algorithmes, démarre sur un réseau de neurones dont les paramètres sont appris. L'initialisation s'effectue en calculant $M_T$ et $M_V$ (19), et en initialisant les mémoires $M_T^{best} = M_T$ et $M_V^{best} = M_V$ et un seuil $Er_{max} = \max\{M_T^{best}, M_V^{best}\}$. Il se déroule ensuite en deux étapes :

Etape 1 : Elimination des neurones cachés

- annuler $w_i^2$ ($i=1\ldots n_1$) et calculer les MAD $M_T(i)$,
- trouver le minimum $M_T(ind) = \min(M_T(i), i=1\ldots n_1)$,
- poser $w_{ind}^2 = 0$ et Réapprendre le réseau,
- recalculer $M_T$ et $M_V$.
- si $M_T \leq Er_{max}(1+\alpha)$ et $M_V \leq Er_{max}(1+\alpha)$ :
  o supprimer le neurone caché ind,
  o $M_T^{best} = \min(M_T^{best}, M_T)$ ; $M_V^{best} = \min(M_V^{best}, M_V)$ ;
  $Er_{max} = \max\{M_T^{best}, M_V^{best}\}$,
- Sinon restaurer les anciens poids et passer à l'étape suivante.

Etape 2 : Elimination des entrées

- annuler $w_{ih}^1$ ($\forall i, h=1\ldots n_0$) et calculer les MAD $M_T(h)$,
- trouver le minimum $M_T(ind) = \min(M_T(h), h=1\ldots n_0)$,
- poser $w_{i,ind}^1 = 0$ ($\forall i$) et Réapprendre le réseau,
- recalculer $M_T$ et $M_V$.
- si $M_T \leq Er_{max}(1+\alpha)$ et $M_V \leq Er_{max}(1+\alpha)$ :
  o supprimer l'entrée ind,
  o $M_T^{best} = \min(M_T^{best}, M_T)$ ; $M_V^{best} = \min(M_V^{best}, M_V)$ ;
  $Er_{max} = \max\{M_T^{best}, M_V^{best}\}$,
- Sinon fin de l'algorithme.

La valeur de $\alpha$ est réglée à 0,025 [28]. Dans la suite de ce papier, cet algorithme sera dénommé « N2PFA ».

## V. LES RESULTATS

*C. Modélisation d'un système statique*

Le système non linéaire à modéliser est tout simplement un perceptron multicouches utilisant 3 entrées et 1 sortie. Le système est décrit par l'équation suivante :

$$y(t) = 1 + \tanh(2.x_1(t) - x_2(t) + 3.x_3(t)) + \tanh(x_2(t) - x_1(t)) + e(t) \quad (20)$$

où $e(t)$ est un bruit blanc additif d'écart type 0,2.

Ce système, supposé inconnu, a été choisi pour éviter les différences entre la forme 'vraie' du système, et celle qui peut être réalisée par le réseau de neurones.

Afin d'effectuer la modélisation, deux jeux de données pour l'apprentissage et la validation comprenant chacun 500 données ont été construits. Ces deux jeux de données comprennent tous les deux 5 variables d'entrée. Toutes les entrées sont des séquences pseudo aléatoires d'amplitude et de durée différentes afin de donner à chaque entrée une influence différente. L'apprentissage initial est réalisé à l'aide d'un réseau de neurones comprenant 5 neurones d'entrée et 8 neurones cachés. L'algorithme d'apprentissage utilisé est l'algorithme de Levenberg Marquardt [23] sur 50000 itérations. 50 jeux de paramètres initiaux ont été obtenus en utilisant une modification de l'algorithme de Nguyen et Widrow [30]. Les 4 algorithmes testés ont tous fonctionné sur les mêmes jeux de paramètres initiaux.

L'ensemble des résultats obtenus sur ce premier système avec les quatre algorithmes testés sont regroupés et synthétisés dans le tableau 1. la première colonne indique le nombre d'entrées conservées par les algorithmes, tandis que la deuxième donne le nombre de neurones cachés conservés et la troisième fournit le nombre de paramètres (poids et biais) constituant le modèle résultant. La dernière colonne indique le temps mis par l'algorithme pour se dérouler. Pour chacune des colonnes, les valeurs minimale maximale et moyenne du paramètre considéré sont relevées pour l'algorithme considéré pour les 50 jeux de paramètres initiaux. Nous trouvons également deux pourcentages. Le premier indique le pourcentage de jeux de poids initiaux fournissant une valeur inférieure à la valeur moyenne, et le deuxième, le pourcentage de jeux de poids initiaux fournissant une valeur supérieure à la valeur moyenne. Chacune des lignes correspond à chacun des algorithmes testés.

|  |  | Nb_I | | Nb_H | | NB_θ | | temps | |
|---|---|---|---|---|---|---|---|---|---|
|  |  | val | % | val | % | val | % | val | % |
| engel | min | 4 | 42% | 5 | 52% | 31 | 40% | 3,1e-2 | 26% |
|  | moy | 4,6 | < > | 7,2 | < > | 47,9 | < > | 4,7e-2 | < > |
|  | max | 5 | 58% | 8 | 48% | 57 | 60% | 6,3e-2 | 74% |
| engel_mod | min | 5 |  | 2 | 46% | 11 | 46% | 0,11 | 50% |
|  | moy | 5 | < > | 3,68 | < > | 24,4 | < > | 0,35 | < > |
|  | max | 5 |  | 6 | 54% | 43 | 54% | 0,61 | 50% |
| N2PFA | min | 4 | 98% | 2 | 56% | 13 | 56% | 1,07 | 52% |
|  | moy | 4,02 | < > | 3,58 | < > | 22,6 | < > | 1,60 | < > |
|  | max | 5 | 2% | 8 | 44% | 49 | 44% | 2,27 | 48% |
| OBS_L_FPE | min | 5 |  | 5 | 2% | 25 | 44% | 13,4 | 56% |
|  | moy | 5 | < > | 7,9 | < > | 50,1 | < > | 20,0 | < > |
|  | max | 5 |  | 8 | 98% | 57 | 56% | 26,1 | 44% |
| OBS_L_MSSE | min | 4 | 80% | 2 | 58% | 8 | 76% | 13,4 | 56% |
|  | moy | 4,2 | < > | 3,78 | < > | 14,4 | < > | 20,0 | < > |
|  | max | 5 | 20% | 8 | 42% | 57 | 24% | 26,1 | 44% |
| OBS_WL_FPE | min | 5 |  | 8 |  | 47 | 32% | 7,56 | 56% |
|  | moy | 5 | < > | 8 | < > | 55,4 | < > | 9,85 | < > |
|  | max | 5 |  | 8 |  | 57 | 68% | 11,9 | 44% |
| OBS_WL_MSSE | min | 4 | 8% | 2 | 18% | 9 | 26% | 7,56 | 56% |
|  | moy | 4,9 | < > | 7,5 | < > | 49,8 | < > | 9,85 | < > |
|  | max | 5 | 92% | 8 | 82% | 57 | 74% | 11,9 | 44% |

Table 1 : résultats sur le système 1

Nous pouvons rappeler que la structure optimale vers laquelle doit tendre nos algorithmes pour le système 1 comprend 3 entrées, 2 neurones cachés et 8 paramètres comme l'indique l'équation (20).

Si nous considérons tout d'abord le nombre d'entrée conservées dans le modèle, nous pouvons immédiatement constaté qu'aucun algorithme ne parvient ne serait ce qu'une fois à ne conserver que les 3 entrées réellement nécessaires. Seul les algorithmes engel, N2PFA et OBS utilisant le critère MSSE sur le jeu de validation avec ou sans apprentissage (OBS_L-MSSE et OBS_WL_MSSE) permettent de supprimer une des deux entrées inutiles. L'étude des pourcentages et des valeurs moyennes permet de constater que l'algorithme N2PFA fournit les meilleurs résultats devant OBS avec réapprentissage (l'absence de réapprentissage dégradant fortement les performances) et devant engel qui supprime une entrée inutile moins d'une fois sur deux (42%).

Lorsque l'on étudie le nombre de neurones cachés conservés dans le modèle, nous obtenons une plus grande dispersion des résultats. Cependant, seul les algorithmes engel_mod, N2PFA, OBS_L_MSSE et OBS_WL_MSSE parviennent à trouver le bon nombre de neurones cachés (2). Les trois algorithmes engel_mod, N2PFA et OBS_L_MSSE fournissent d'ailleurs des résultats très comparables ce qui est dénoté par leurs valeurs moyennes (3,68 ; 3,58 ; 3,78 respectivement), les pourcentages de résultats inférieurs aux valeurs moyennes (46% ; 56% ; 58% respectivement) et les valeurs maximales (6 ; 8 ; 8 respectivement). L'absence de réapprentissage dégrade fortement les performances de l'algorithme OBS.

Si l'on s'en tient à ces informations, l'algorithme N2PFA fournit les meilleurs résultats. Cette affirmation est nuancée lorsque l'on étudie le nombre de paramètres constituant les modèles résultants. La première remarque que l'on peut faire est qu'il semble que seul l'algorithme OBS_L_MSSE (et OBS_WL_MSSE) permet de trouver les 8 paramètres de la structure optimale. Malheureusement, lorsque l'on étudie la structure concernée, on s'aperçoit qu'un certain nombre de paramètres n'appartenant pas normalement à la structure sont conservés au détriment de paramètres qui aurait du être conservés (en particulier, au moins une entrée inutile a été conservée). Au contraire de cet algorithme, les algorithmes engel_mod et N2PFA parviennent à approcher assez bien la structure optimale sans supprimer de connexions utiles en conservant au minimum 11 et 13 paramètres respectivement. L'étude des pourcentages et des valeurs moyennes et maximales permet de constater que ces deux algorithmes fournissent des résultats très similaires.

Enfin, lorsque l'on étudie les temps de calcul, on constate que l'algorithme OBS est beaucoup plus long que ses trois concurrents même quand aucun réapprentissage n'est demandé. L'algorithme engel_mod est lui bien plus rapide que son principal concurrent N2PFA avec un rapport moyen de 4 entre le temps de calcul de engel_mod, 0,35 et celui de N2PFA 1,60. Cette différence est principalement due au réapprentissage nécessité par N2PFA et pas par engel_mod.

### D. Modélisation d'un système dynamique

Le deuxième système est à nouveau un perceptron multicouches utilisant cette fois ci des entrées retardées. Le système est décrit par l'équation suivante :

$$y(t) = 1 + \tanh(x_1(t-2) - x_2(t) + 3.x_2(t-1)) \\ + \tanh(x_1(t-2) - x_2(t-2)) + e(t) \qquad (21)$$

où $e(t)$ est un bruit blanc additif d'écart type 0,2.

|  |  | Nb_I | | Nb_H | | NB_θ | | temps | |
|---|---|---|---|---|---|---|---|---|---|
|  |  | val | % | val | % | val | % | val | % |
| engel | min | 8 | 14% | 7 | 38% | 71 | 42% | 3,1e-2 | 66% |
|  | moy | 9,84 | < > | 9,44 | < > | 112,9 | < > | 5,0e-2 | < > |
|  | max | 10 | 89% | 10 | 62% | 121 | 58% | 9,4e-2 | 34% |
| engel_mod | min | 10 |  | 2 | 70% | 22 | 48% | 0,22 | 62% |
|  | moy | 10 | < > | 3,14 | < > | 36,8 | < > | 0,70 | < > |
|  | max | 10 |  | 6 | 30% | 64 | 52% | 1,03 | 38% |
| N2PFA | min | 4 | 70% | 2 | 52% | 13 | 60% | 3,49 | 48% |
|  | moy | 5,52 | < > | 3,92 | < > | 32,7 | < > | 5,74 | < > |
|  | max | 10 | 30% | 10 | 48% | 97 | 40% | 8,11 | 52% |
| OBS_L_MSSE | min | 4 | 48% | 2 | 46% | 8 | 52% | 142,1 | 52% |
|  | moy | 5,72 | < > | 4,7 | < > | 17 | < > | 160,4 | < > |
|  | max | 9 | 52% | 9 | 54% | 32 | 46% | 180,0 | 48% |

Table 2 : résultats sur le système 2

L'entrée $x_1$ ($x_2$ resp.) est une séquence pseudo aléatoire d'amplitude variant entre -1 et 1 (0 et 1,5 resp.) et de durée variant entre 5 et 10 (8 et 15 resp.). Le vecteur d'entrée utilisé lors de l'apprentissage est constitué des deux entrées $x_1$ et $x_2$ et de leurs retards respectifs t, t-1, t-2, t-3 et t-4 ce qui fait 10 neurones d'entrée. Le réseau initial possède 10 neurones cachés. L'apprentissage initial a été effectué sur 50 jeux de paramètres initiaux en utilisant 50000 itérations.

L'ensemble des résultats obtenus sur ce système 2 sont regroupés et synthétisés dans le tableau 2. Nous pouvons rappeler que la structure optimale vers laquelle doit tendre nos algorithmes pour le système 2 comprend 4 entrées, 2 neurones cachés et 8 paramètres comme l'indique l'équation (21). Une étude rapide de ces résultats permet de constater qu'à nouveau les algorithmes engel_mod et N2PFA fournissent les meilleurs résultats (OBS_L_MSSE supprimant des paramètres utiles).

En particuliers, ces deux algorithmes sont les seuls à fournir les 2 neurones cachés utiles et conservent un nombre de paramètres relativement similaire. Cependant, si N2PFA parvient à fournir le bon nombre d'entrées, il le fait dans un temps 8 fois plus long en moyenne que engel_mod, cela étant du encore une fois au réapprentissage nécessaire.

## VI. CONCLUSION

Dans cet article, nous avons présenté un nouvel algorithme pour déterminer la structure optimale d'un perceptron multicouches. Cet algorithme a été comparé à trois autres et les résultats montrent bien l'intérêt de l'algorithme proposé tant en ce qui concerne la structure trouvée que le temps de calcul nécessaire. Dans nos travaux futurs, nous allons tester cet algorithme sur des cas réels d'applications et notamment, dans le cadre de la réduction de modèles de simulations d'ateliers.